\documentclass{article}

\usepackage{microtype}
\usepackage{graphicx}
\usepackage{subcaption}
\usepackage{booktabs} 
\usepackage{array} 

\usepackage[english]{babel}

\usepackage{hyperref}

\usepackage{algorithm}
\usepackage{algpseudocode}
\makeatletter
\@namedef{ver@algorithmic.sty}{9999/12/31}
\makeatother

\usepackage[preprint]{icml2027}

\usepackage{amsmath,amsfonts,bm}

\def\eqref#1{equation~\ref{#1}}

\def\1{\bm{1}}

\DeclareMathAlphabet{\mathsfit}{\encodingdefault}{\sfdefault}{m}{sl}
\SetMathAlphabet{\mathsfit}{bold}{\encodingdefault}{\sfdefault}{bx}{n}

\usepackage{amssymb}
\usepackage{mathtools}
\usepackage{amsthm}
\usepackage{url}

\usepackage{enumitem}
\usepackage{wrapfig}
\usepackage{needspace}
\usepackage{tikz} 
\graphicspath{{figures/}}
\usepackage{cleveref}

\theoremstyle{plain}

\theoremstyle{definition}

\theoremstyle{remark}

\usepackage[textsize=tiny]{todonotes}

\icmltitlerunning{Learning to Reason in Real-Time}

\begin{document}

\twocolumn[
  \icmltitle{Learning to Reason in Real-Time}



  \icmlsetsymbol{equal}{*}

  \begin{icmlauthorlist}
    \icmlauthor{Xavier Schneider}{equal,uw}
    \icmlauthor{Benjamin Schneider}{equal,uw}
    \icmlauthor{Sun Sun}{nrc}
    \icmlauthor{Olga Vechtomova}{uw}
    \icmlauthor{Victor Zhong}{uw}
  \end{icmlauthorlist}

  \icmlaffiliation{uw}{University of Waterloo, Waterloo, Ontario, Canada}
  \icmlaffiliation{nrc}{National Research Council Canada}

  \icmlcorrespondingauthor{Xavier Schneider}{xschneider@uwaterloo.ca}

  \icmlkeywords{Machine Learning, ICML}

  \vskip 0.3in
]



\printAffiliationsAndNotice{\icmlEqualContribution}

\begin{abstract}
AI agents that operate in the real world must reason correctly and on-time, yet they currently do not account for how time should alter their reasoning.
Modern agents learn to do tasks in environments that are entirely \emph{synchronous}: where the world waits for the agent to act before changing.
A \emph{real-time} environment, by contrast, changes independent of the agent.
Existing approaches try to bridge this gap by imposing a ratio of thinking to acting. 
However, for an agent to succeed in real-time it must learn to act quickly when delay is costly and to reason longer when the benefit outweighs the missed opportunities. 
To this end, we introduce \textbf{real-time reasoning training (RRT)} as the class of training techniques for real-time reasoning agents.
We develop three standard RRT techniques: (1) using real-time data, (2) decoupling acting and observing, and (3) curriculum learning in time.
Together, these simple techniques substantially outperform strong baselines. 
Across time pressures, we outperform the strongest non-RRT baseline by an average of 33\% reward on in-distribution and 55\% reward on out-of-distribution tasks in two real-time environments: Frogger and Snake.
These results demonstrate that we can teach agents to balance reasoning and responsiveness under real-time constraints, and that RRT opens a broad, new set of capabilities for real-time agents.
\end{abstract}

\section{Introduction}
\label{sec:intro}

For Large Language Model (LLM) agents, time is frozen. 
Such agents are trained on tasks like coding, computer-use and mathematics, where the world waits for the agent to act before changing. 
But in the real-world, time waits for no one.
Consequently, people learn to balance thinking and acting.
We run to catch a bus, we carefully plan important life decisions and we use all available time to answer the hardest question on an exam. 
This ability to reason in real-time is inseparable from our ability to successfully operate in the world.

Despite substantial effort to improve agent performance with LLM reasoning, current agents perform poorly in even simple real-time environments. 
\citet{wen2026realtime} show that LLM agents struggle to play simple games under even mild time pressure. 
This is because these agents are trained in environments that are entirely synchronous: where the environment waits for the agent to act before updating.
Such policies transfer poorly to real-time environments, which update according to their own schedule.

Previous work has attempted to address this limitation by deploying LLM-agents in a multi-level scaffold.
Wen et al.~\cite{wen2026realtime} propose using a faster secondary LLM to infer an action from the primary LLM's reasoning trace, allowing the agent to take actions while still reasoning.
However, these methods rely on a predefined ratio of thinking to acting.
\emph{In the real-time setting, an agent should act quickly when delay is costly and reason longer when the benefit outweighs the missed opportunities.} 
Since the optimal amount of reasoning changes depending on the task and circumstances of the agent, the agent itself must be able to choose when to reason more, and when to act.

To this end, we introduce \textbf{real-time reasoning training (RRT)} as the class of training techniques for real-time reasoning agents.
We consider RRT to be when a reasoning agent is trained under real-time constraints, enabling the developement of real-time strategies.
We develop three core RRT techniques: 
(1) Using real-time data: for an agent to be successful in real-time environments, it must be trained using real-time interactions, not synchronous ones.
(2) Decoupling acting and observing: we add observations to context as they are created by the environment, instead of only after an action (as in the synchronous case).
This allows the agent to learn to incorporate new information into its decision making.
(3) Curriculum Learning in Time: We begin by training our agent in slow real-time environments, and gradually increase the time pressure. 
This allows the agent to adjust to harsher time pressure. 

We evaluate our RRT techniques in Frogger and Snake.
In each environment, we evaluate on held-out seeds of three in-distribution tasks, used during training, and on one out-of-distribution task over a range of time pressures. 
We compare against synchronous, non-reasoning, scaffolded and naive RRT agents (\Cref{exp:main}). 
Combined, our techniques are on-par or exceed all other methods on held-out seeds in every task at every tested speed; outperforming the strongest non-RRT baseline by an average of 33\% and 55\% reward on in-distribution and out-of-distribution tasks, respectively.
To validate the contribution of each of our RRT techniques, we perform an ablation in \Cref{exp:ablation}. 
%
%
Finally, to quantify what advantages our techniques have over other methods, we perform a detailed error analysis (\Cref{exp:fail}). 
%

To summarize, our work makes three key contributions:

\begin{enumerate}[
    leftmargin=1.2em,
    itemsep=2pt,
    topsep=1pt,
    parsep=0pt,
    partopsep=0pt
]
\item We introduce \emph{real-time reasoning training (RRT)} as the class of training techniques for real-time reasoning agents.
\item We develop three core RRT techniques: Using real-time reasoning trajectories, decoupling acting and observing and curriculum learning in time.
\item We provide experimental evidence showing the benefit of our techniques over scaffolded and naive RRT methods.

\end{enumerate}

\section{Real-Time Environments}
\label{sec:background}

In conventional reasoning training, the environment is \emph{synchronous}: while the LLM agent produces its reasoning trace the environment is static waiting for an action or tool call~\citep{feng2026retool}.
This paradigm works well for coding~\citep{qwen38} and most computer-use~\citep{kimiteam2026kimik25visualagentic}, where the environment is naturally static without interaction.
By contrast, \emph{real-time} environments are dynamic: they change while the agent reasons and not only when the agent acts.
In this case, actions taken by the agent are executed against the environment at the moment they are produced.

\textbf{Notation.} We model an environment as a tuple $\mathcal{E}=(\mathcal{S},\mathcal{A},\mathcal{T})$, where $\mathcal{S}$ is the state
space, $\mathcal{A}$ is the action space, and
$\mathcal{T}:\mathcal{S}\times\mathcal{A}\rightarrow\mathcal{S}$ is the state transition
function. We assume an initial state $s_0$ and we let
$\mathcal{S}_{\mathrm{term}}\subseteq\mathcal{S}$ denote the terminal states.
We represent an agent trajectory as an ordered collection $H_T = \langle s_0, r_0, a_0, \ldots, s_{T-1}, r_{T-1}, a_{T-1}, s_T \rangle $. 
At timestep $t$, $s_t$ is the environment state, $r_t$ denotes the reasoning tokens produced by the agent between subsequent states and $a_t$ is the action taken by the agent.
We use $\oplus$ to denote appending elements to the end of a trajectory, i.e., $H_{t+1} = H_t \oplus \langle r_t, a_t, s_{t+1} \rangle$.
$C_t$ is the agent's context, built from its trajectory, $C_t = \textsc{Context}(H_t)$.
We use $(\hat r_t, \hat a_t) \sim \pi(\,\cdot \mid C_t)$ to indicate asynchronously sampling reasoning and an action from an agent's policy $\pi$, conditioned on its context.

\begingroup
\setlength{\intextsep}{7pt}
\begin{algorithm}[H]
\caption{Synchronous Environment}
\label{alg:turnbased}
\begin{hyphenrules}{nohyphenation}
\small
\begin{algorithmic}[1]
\setlength{\itemsep}{.2ex}
\setlength{\parskip}{0pt}
\setlength{\parsep}{0pt}

\Require $\mathcal{E}=(\mathcal{S},\mathcal{A},\mathcal{T})$,\, $\pi$,\, $s_0$,\, $\mathcal{S}_{\mathrm{term}}$

\State $t \leftarrow 0$
\State $H_0 \leftarrow \langle s_0 \rangle$
\While{$s_t \notin \mathcal{S}_{\mathrm{term}}$}
  \State $r_t, a_t \sim \textbf{await}\ \pi(\,\cdot \mid \textsc{Context}(H_t))$ \label{line:sample}
  \State $s_{t+1} \leftarrow \mathcal{T}(s_t,a_t)$ \label{line:transition}
  \State $H_{t+1} \leftarrow H_t \oplus \langle r_t, a_t, s_{t+1} \rangle$
  \State $t \leftarrow t+1$
\EndWhile
\end{algorithmic}
\normalsize
\end{hyphenrules}
\end{algorithm}
\vspace{-10pt}
\endgroup

\textbf{Synchronous Environments.} \Cref{alg:turnbased} shows the synchronous agent loop. 
Given $s_t$, the policy may take arbitrarily long to sample $r_t, a_t\sim\pi$ conditioned on its context.
line~\ref{line:sample} shows that the environment waits for the asynchronous agent sampling to complete before proceeding.
Only after the sample is complete does the environment transition to
$s_{t+1}$.
Because sampling on line~\ref{line:sample} runs to completion before the
transition on line~\ref{line:transition},
the agent always has time to complete its reasoning before acting.

\begingroup
\setlength{\intextsep}{7pt}
\begin{algorithm}[H]
\caption{Real-Time Environment}
\label{alg:realtime}
\begin{hyphenrules}{nohyphenation}
\small
\begin{algorithmic}[1]
\setlength{\itemsep}{.2ex}
\setlength{\parskip}{0pt}
\setlength{\parsep}{0pt}

\Require $\mathcal{E}=(\mathcal{S},\mathcal{A},\mathcal{T})$,\, $\pi$,\, $s_0$,\, $\mathcal{S}_{\mathrm{term}}$,\, $\Delta_{\mathcal{E}}$

\State $t \leftarrow 0$
\State $H_0 \leftarrow \langle s_0 \rangle$
\While{$s_t \notin \mathcal{S}_{\mathrm{term}}$}
  \State $\hat r_t, \hat a_t \sim \pi(\,\cdot \mid \textsc{Context}(H_t))$ \label{line:rt-sample}
  \State $\textsc{Wait}(\Delta_{\mathcal{E}})$
  \If{$\textsc{IsReady}(\hat a_t)$}
    \State $a_t \leftarrow \hat a_t$
  \Else
    \State $a_t \gets \texttt{[NULL]}$
  \EndIf
  \State $s_{t+1} \leftarrow \mathcal{T}(s_t,a_t)$
  \State $H_{t+1} \leftarrow H_t \oplus \langle \hat r_t, a_t, s_{t+1} \rangle$
  \State $t \leftarrow t+1$
\EndWhile
\end{algorithmic}
\normalsize
\end{hyphenrules}
\end{algorithm}
\vspace{-10pt}
\endgroup

\textbf{Real-Time Environments.}
\Cref{alg:realtime} shows the real-time agent loop.
Unlike in a synchronous environment, a real-time environment has an \emph{update interval}
$\Delta_{\mathcal{E}}>0$ which determines how frequently the environment transitions, independent of policy sampling.
The lower the value of $\Delta_{\mathcal{E}}$, the more \emph{time pressure} the agent experiences.  
At state $s_t$, the agent's policy is sampled asynchronously to
produce reasoning tokens $\hat r_t$ and a candidate action $\hat a_t$.
If sampling finishes within the current interval, the environment applies $a_t=\hat a_t$.
Otherwise, it applies the \texttt{[NULL]} action: representing the environment transitioning without the agent producing an action.
The environment then transitions to $s_{t+1}=\mathcal{T}(s_t,a_t)$.
We note that given a sufficiently high $\Delta_{\mathcal{E}}$, \Cref{alg:realtime} becomes
\Cref{alg:turnbased}, i.e. it becomes synchronous.

\section{Method}
\label{sec:method}

To adapt LLM agents to real-time environments, we develop three core RRT techniques:
\textbf{(1)} Trajectories have to be collected from a real-time environment, not a synchronous one.
\textbf{(2)} The agent's reasoning needs to contain states that were produced during reasoning, to prevent the policy from reasoning using information from stale states.
\textbf{(3)} The time pressure should be annealed from low to high during training. This allows the reasoning policy to adapt strategies learned under low time pressure to more demanding settings, resulting in a more robust final policy.

\subsection{Collecting real-time trajectories}
Success in real-time environments depends on an agent's ability to allocate reasoning in a time-aware way. 
Intuitively, an agent should respond quickly when delay is costly and reason longer when the benefit outweighs the missed opportunities.

\begin{figure}[h]
\vspace{-2ex}
\centering
\includegraphics[width=\columnwidth]{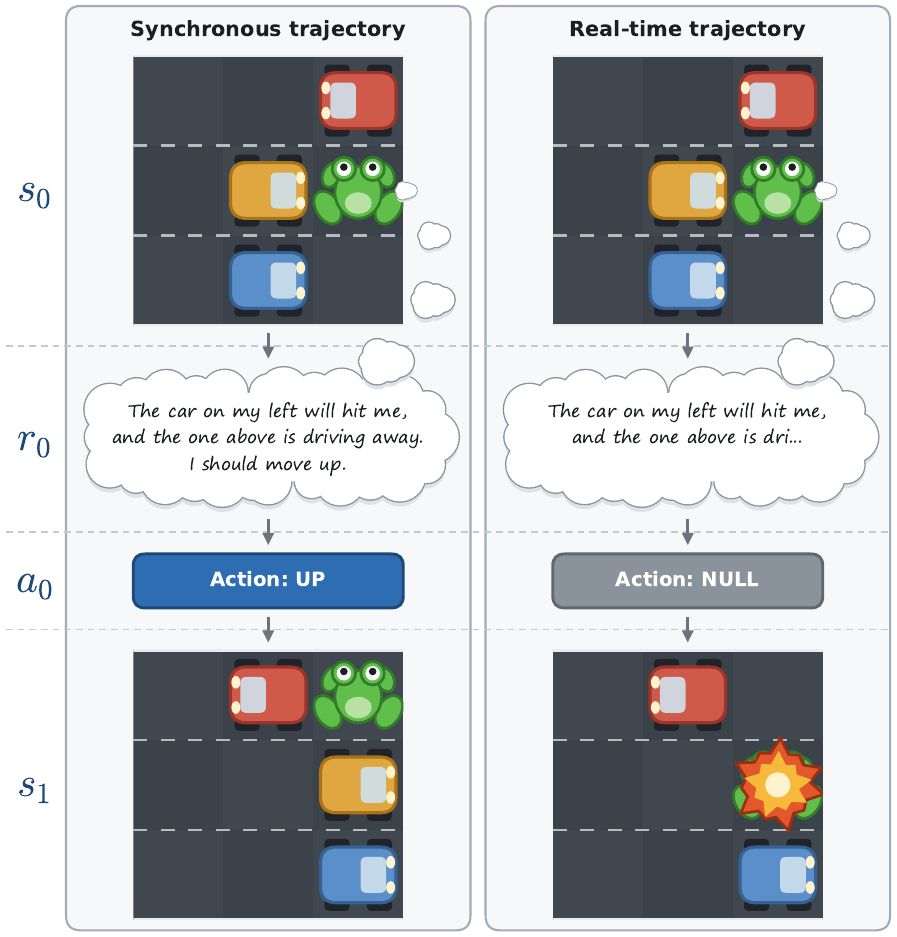}
\caption{Synchronous vs.\ real-time trajectories in Frogger. \textbf{Left:} in a synchronous environment the world waits
for the agent to finish its reasoning, so it moves up
and survives. \textbf{Right:} in a real-time environment the
environment advances after $\Delta_{\mathcal{E}}$ tokens, cutting the same reasoning off before an action is produced.}
\label{fig:frogger-comparison}
\vspace{-2ex}
\end{figure}

Therefore, we collect agent trajectories using real-time environments (\Cref{alg:realtime}), instead of the traditional synchronous environment (\Cref{alg:turnbased}) that reasoning agents are trained in.
We model time pressure ($\Delta_{\mathcal{E}}$) as \emph{tokens generated}, which is a controlled, hardware independent setup that correlates closely with wall-clock time~\citep{wen2026realtime}.
In a trajectory collected in a synchronous environment, each action is chosen by the policy using the most recent observation.
However, in a real-time environment, it is possible that $a_i = \texttt{[NULL]}$, corresponding to the case where the reasoning policy did not produce an action before the environment transition.

\subsection{Decoupling acting and observing}

In synchronous reasoning training, acting and observing are \emph{coupled}: observations are only added to the agent's context after each action.
%
%
However, if the state of the underlying environment has changed, as in a real-time environment, the chosen action may no longer be coherent.
This \emph{stale state problem}--where the agent reasons using outdated information--is a failure mode introduced when naively adapting synchronous reasoning training to real-time environments.

\begin{figure}[h]
\vspace{-2ex}
\centering
\includegraphics[width=\columnwidth]{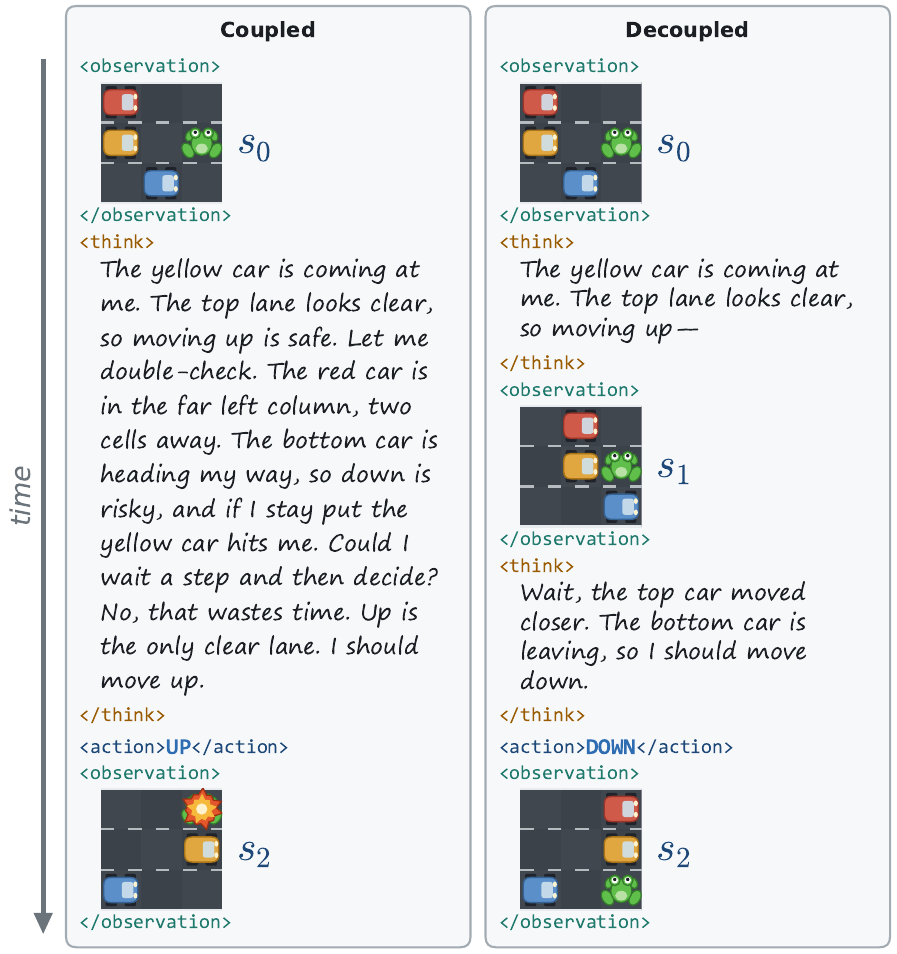}
\caption{Coupled vs.\ decoupled observation in Frogger.
\textbf{Left:} with acting and observing coupled, no new observation arrives while the agent reasons. \textbf{Right:} with decoupling, the new observation is added to the agent's context mid-reasoning, allowing it to update its ongoing reasoning to avoid a collision.}
\label{fig:frogger-history}
\vspace{-2ex}
\end{figure}

To address this issue, we decouple acting and observing. Rather than only adding state to agent context after an action, we add state to context as it is created by the environment.
This allows the agent to update its ongoing reasoning and choose a different action using the most up-to-date environment information.
This prevents the observed failure mode (\Cref{exp:ablation}), where the agent makes a poor decision due to reasoning with outdated environment state.

More formally, given the agent trajectory at timestep $t$:
\begin{equation}
H_t = \langle s_0, r_0, a_0, \ldots, s_{t-1}, r_{t-1}, a_{t-1}, s_t \rangle .
\end{equation}
Let $\mathcal{N}_t = \{\, i < t : a_i = \texttt{[NULL]} \,\}$ be the set of timesteps before $t$ at which the action was $\texttt{[NULL]}$.

With coupled observations, the agent's context at timestep $t$ is
\begin{equation}
C_t = H_t \setminus \{\, a_i,\ s_{i+1} : i \in \mathcal{N}_t \,\},
\end{equation}
where $H_t \setminus X$ denotes the subsequence of $H_t$ obtained by removing the elements of $X$, with the remaining elements kept in their original order.

Conversely, when decoupled, the agent's context retains the observation even when the action is $\texttt{[NULL]}$:
\begin{equation}
C_t = H_t \setminus \{\, a_i : i \in \mathcal{N}_t \,\}.
\end{equation}

%

\subsection{Curriculum learning in time}

We find that learning exclusively under high time pressure results in a poor final policy~(\Cref{exp:ablation}).
Instead, we adopt a curriculum learning approach: throughout training we increase the time pressure on a fixed schedule.
This allows the agent to experiment, develop, and refine its policy before experiencing harsh time-constraints.
The agent therefore continually adapts a well-rounded policy for increasingly demanding real-time settings, rather than learning crutches and brittle compensatory strategies.

\begin{figure}[h]
\vspace{-2ex}
\centering
\includegraphics[width=\columnwidth]{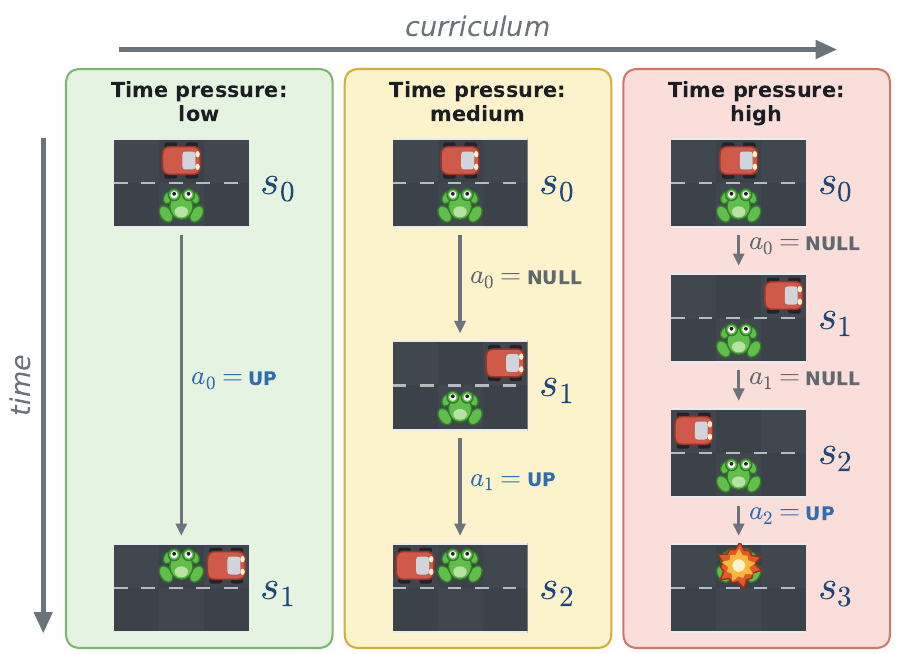}
\caption{Increasing time pressure on a schedule in Frogger. As time pressure increases (i.e. $\Delta_{\mathcal{E}}$ \emph{is decreased}), the agent is forced to adapt its policy to make faster decisions to survive.}
\label{fig:frogger-curriculum}
\vspace{-1ex}
\end{figure}

\begin{figure*}[t]
\centering
\includegraphics[width=0.8\textwidth]{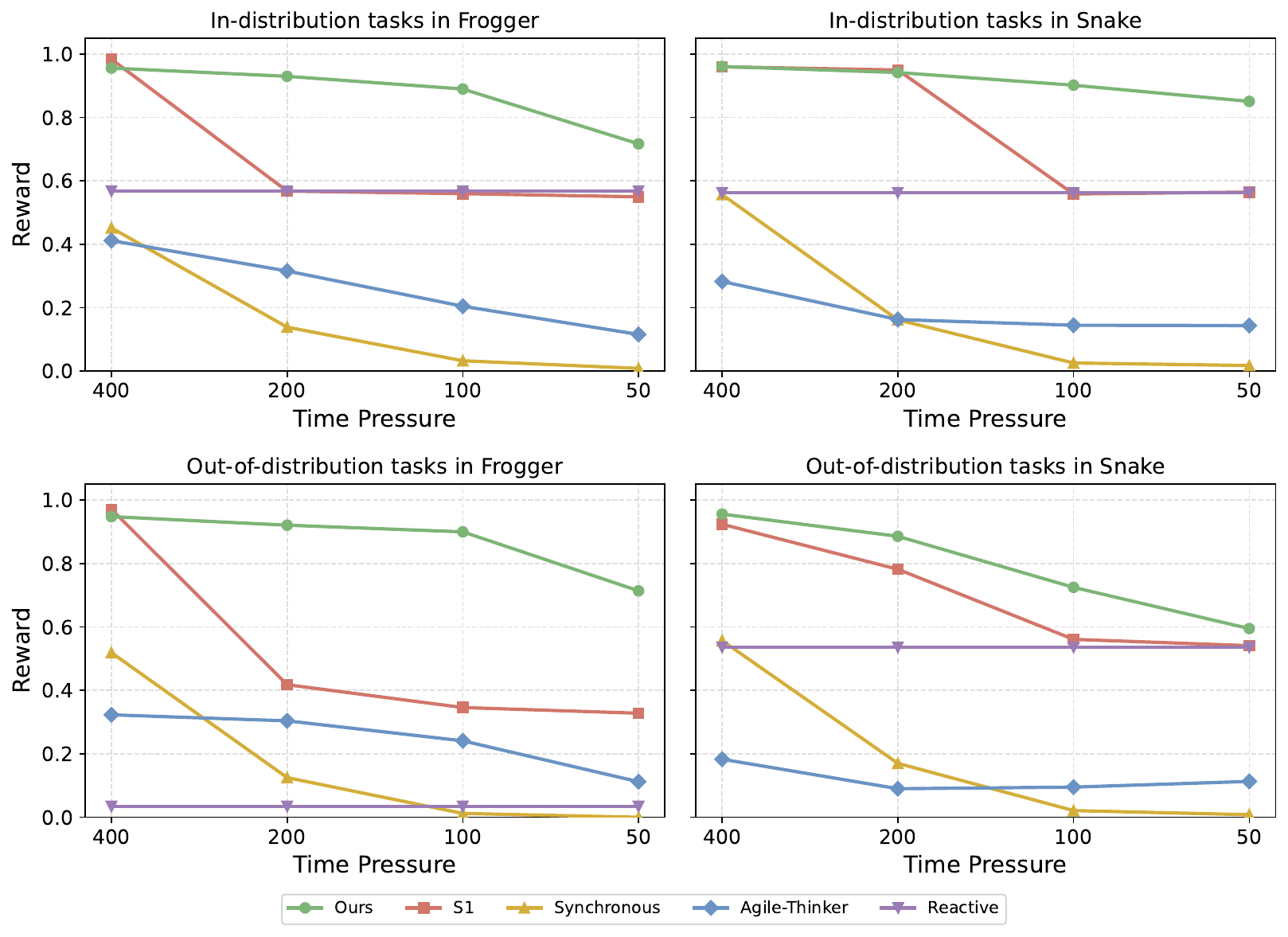}
\caption{Performance on real-time tasks. Earned reward in Frogger and Snake as time pressure increases (i.e. $\Delta_{\mathcal{E}}$ \emph{is decreased}), on the in-distribution tasks (\textbf{top}) and the out-of-distribution task (\textbf{bottom}). Our method is compared with the Synchronous, S1, Reactive, and AgileThinker baselines. Each point is the average over 256 held-out seeds with 8 rollouts each.}
\label{fig:main-exp}
\end{figure*}

We design our schedule to move from low to high time pressure at set intervals, where the change is proportional to the time pressure at the previous interval.
%
%
Let $\Delta_{\mathcal{E}}^{start}$ and $\Delta_{\mathcal{E}}^{end}$ be the starting and target environment update intervals, $\kappa$ the decay interval, and $N$ the total number of training steps.
The environment update interval at step $n$ (denoted $\Delta_{\mathcal{E}}^{n}$) is:

\begin{equation}
\Delta_{\mathcal{E}}^{n} = \Delta_{\mathcal{E}}^{start} \cdot \rho^{\lfloor{} \frac{n}{\kappa} \rfloor{}},
\qquad
\rho = \left(\frac{\Delta_{\mathcal{E}}^{end}}{\Delta_{\mathcal{E}}^{start}}\right)^{\frac{\kappa}{N}}
\end{equation}

Here $\rho$ is the \emph{proportional decrease in the environment update interval} ($\Delta_{\mathcal{E}}$) each $\kappa$ training steps, computed from the fixed values of $\Delta_{\mathcal{E}}^{start}$, $\Delta_{\mathcal{E}}^{end}$, $\kappa$ and $N$.
The resulting schedule is therefore an exponential curve, discretized into intervals of length $\kappa$, that reaches $\Delta_{\mathcal{E}}^{end}$ at step $N$.

\section{Experiments}
\label{sec:experiments}

We evaluate our RRT techniques in two real-time game environments: Frogger and Snake.
These games are designed to balance planning (like deciding a route through busy traffic) with real-time reactions (such as avoiding an oncoming car).
For each environment, we construct four task variants, each with their own objective and rewards.
Each task is seeded, resulting in a large number of distinct configurations per task.
We use different seeds for training and evaluation.
Furthermore, we randomly select one task per environment to act as an \emph{out-of-distribution task}---a task that is not used during training---to demonstrate that our techniques can generalize across tasks, not just environment configurations.
We provide a detailed description of environments, tasks, and rewards in~\Cref{apx:evaluation-tasks}. 
Through our experiments, we investigate the following:
\begin{itemize}[
    leftmargin=1.2em,
    itemsep=1.5pt,
    topsep=1pt,
    parsep=0pt,
    partopsep=0pt
]
    \item How do our RRT techniques compare to baselines on in-distribution and out-of-distribution tasks?~(\Cref{exp:main})
    \item How do using real-time trajectories, decoupling acting and observing, and curriculum learning in time each contribute to real-time performance?~(\Cref{exp:ablation})
    \item What is the source of the advantage of our method over baselines?~(\Cref{exp:fail})
\end{itemize}

\textbf{Training settings and implementation details.}
We use Qwen3-4B~\cite{yang2025qwen3technicalreport} as our starting LLM policy.
Each run is trained using GRPO~\citep{shao2024deepseekmathpushinglimitsmathematical}. Following \citet{yu2025dapoopensourcellmreinforcement}, we train with no KL divergence penalty and apply asymmetric clipping ($0.28$ high, $0.2$ low) on PPO updates.
We train for $N=640$ steps, with a batch and group size of $256$ and $8$ rollouts, respectively.
We sample rollouts using a temperature of $1.0$.
Inference and training both use bfloat16 floating point precision. 
For our schedule, we set $\Delta_{\mathcal{E}}^{start}=1600$ and $\Delta_{\mathcal{E}}^{end}=\Delta_{\mathcal{E}}$, for all training runs.
We decay the schedule every $\kappa=40$ steps.
Experiments were run using $2\times$ H200 GPUs.
%

\textbf{Baselines.}
As with our method, all baselines adopt Qwen3-4B~\cite{yang2025qwen3technicalreport} as the initial reasoning policy and the same training settings. We compare our method with the
following approaches:
\textbf{Synchronous}~\citep{Guo_2025}:
We employ reasoning training on an LLM policy in synchronous versions of our environments. This agent learns to complete tasks without real-time constraints.
\textbf{S1}~\citep{muennighoff2025s1simpletesttimescaling}: We create a naive RRT baseline by training on top of S1 budget forcing. During training, reasoning is extended using the phrase ``wait'' and is truncated by forcing an action output when it reaches the budget cap. We set the budget to be $\Delta_{\mathcal{E}}$---the maximum number of tokens before the environment transitions.
\textbf{Reactive}: We train a baseline that only outputs actions, not reasoning. This results in a policy that acts almost instantly, but cannot use reasoning to improve its decision-making.
\textbf{AgileThinker}~\citep{wen2026realtime}: AgileThinker adapts a synchronous LLM agent into a real-time agent via a dual-thread system: one thread reasons continually while another thread uses that reasoning as input to make fast decisions. We use the Synchronous baseline to initialize AgileThinker with a strong policy that understands each environment.

\subsection{Performance on Real-Time Tasks}
\label{exp:main}
We evaluate each method in every environment on in-distribution and out-of-distribution tasks over a range of time pressures ($\Delta_{\mathcal{E}}$) and report our results in \cref{fig:main-exp}. 
A real-time environment can become synchronous if $\Delta_{\mathcal{E}}$ is too large (\cref{sec:background}), so we evaluate over a range of $\Delta_{\mathcal{E}}$ where we observe a tradeoff between reasoning and acting for all methods. 
We find that:

\textbf{Our method is the most robust to time pressure.}
At low time pressure ($\Delta_{\mathcal{E}}=400$), our method is on par with the strongest baseline, S1, on both in-distribution and out-of-distribution tasks.
However, as time pressure increases to $\Delta_{\mathcal{E}}=50$, S1 loses 43\% of its reward on average on in-distribution tasks and 54\% on out-of-distribution tasks, while our method loses just 18\% and 31\%, respectively.
The Synchronous baseline has the sharpest decline, losing nearly all ($\geq 97\%$) of its reward at high time pressure on all tasks.
AgileThinker trades some of the performance of the Synchronous baseline under low time pressure for better performance at high time pressure, losing 60\% and 52\% on in-distribution and out-of-distribution tasks, but falls significantly behind our method and S1 in terms of absolute performance.
The Reactive baseline is not sensitive to time pressure since it does not perform reasoning, but it achieves low reward overall on both in-distribution and out-of-distribution tasks.

\textbf{Agents should learn under real-time constraints.}
The two approaches that train reasoning under real-time constraints---our method and S1---match or exceed every other baseline in all settings, achieving an average of 83\% and 61\% reward on out-of-distribution tasks over all time pressures.
Non-RRT methods like AgileThinker and the Synchronous baseline perform poorly even under mild time pressure, each achieving an average of just 18\% on out-of-distribution tasks.
These methods are based on a policy that was trained in synchronous environments without real-time constraints, and they have not developed the real-time capabilities of RRT methods.

\textbf{Reasoning is always useful.}
While the performance of all reasoning methods decreases as time pressure increases, we find that even at extreme time pressure RRT methods match or beat the non-reasoning Reactive baseline.
On in-distribution tasks, our method and S1 beat the Reactive baseline with average margins of 33\% and 15\%, respectively.
We also observe a sharp drop in performance for the Reactive baseline on out-of-distribution tasks, with average reward dropping by 28\% compared to in-distribution tasks.
The reasoning method with the largest drop was S1, which drops by just 10\%.
This suggests that without reasoning the Reactive baseline is also worse at adapting to unseen tasks.

\subsection{Ablation}
\label{exp:ablation}

In \cref{tab:ablation} we ablate across our RRT techniques and measure their contribution to real-time performance.
We evaluate on in-distribution and out-of-distribution Snake tasks at different time pressures.

\begin{table}[h]
\centering
\footnotesize
\setlength{\tabcolsep}{3pt}
\resizebox{\columnwidth}{!}{%
\begin{tabular}{l ccc ccc}
\toprule
& \multicolumn{3}{c}{In-distribution} & \multicolumn{3}{c}{Out-of-distribution} \\
\cmidrule(lr){2-4} \cmidrule(lr){5-7}
$\Delta_{\mathcal{E}}$ & 200 & 100 & 50 & 200 & 100 & 50 \\
\midrule
Synchronous              & 0.16 & 0.03 & 0.02 & 0.17 & 0.02 & 0.01 \\
+ Real-time trajectories          & 0.56 & 0.54 & 0.55 & 0.57 & 0.56 & 0.52 \\
+ Decoupling acting \& observing  & 0.94 & 0.57 & 0.60 & 0.67 & 0.56 & 0.57 \\
+ Curriculum learning in time     & \textbf{0.94} & \textbf{0.90} & \textbf{0.85} & \textbf{0.89} & \textbf{0.73} & \textbf{0.60} \\
\bottomrule
\end{tabular}}
\caption{Additive ablation on Snake. Each row adds one component of our method on top of the row above, starting from the Synchronous baseline. We report earned reward on the in-distribution tasks and on the out-of-distribution task under three levels of time pressure; smaller $\Delta_{\mathcal{E}}$ means more pressure. Averaged over 256 held-out seeds, 8 rollouts each.}
\label{tab:ablation}
\end{table}

As was the case previously, the Synchronous baseline performs poorly in all settings, and is particularly bad at the highest time pressures.
Adding real-time trajectory training results in an average gain of 48 points in task performance (both in-distribution and out-of-distribution), with a relatively uniform increase in performance at different time pressures: 40 points at $\Delta_{\mathcal{E}}=200$ vs 52 points at $\Delta_{\mathcal{E}}=50$.
By decoupling acting and observing, performance increases by 10 points over all settings, but unlike adding real-time trajectories, the gain is skewed towards lower time pressures with 24 points at $\Delta_{\mathcal{E}}=200$ compared to 5 points at $\Delta_{\mathcal{E}}=50$.
Finally, Curriculum learning in time adds an average of 17 points over all settings, this time with most of the effect skewed towards higher time pressures, 25 at $\Delta_{\mathcal{E}}=100$ and 14 at $\Delta_{\mathcal{E}}=50$ compared to 11 at $\Delta_{\mathcal{E}}=200$.
Thus, together our RRT techniques contribute an average of 75 points over all settings, with the gain being relatively uniform: 75 points at $\Delta_{\mathcal{E}}=200$ compared to 71 points at $\Delta_{\mathcal{E}}=50$.


\subsection{Failure Analysis}
\label{exp:fail}

\Cref{exp:main} shows that our method is more robust to time pressure than baselines, but it does not explain \emph{why} it has this advantage. 
To this end, we perform a large-scale failure analysis on training and deployment trajectories.
In the real-time setting, we identify two primary categories of error:

\begin{enumerate}[
    leftmargin=1.2em,
    itemsep=1.5pt,
    topsep=1pt,
    parsep=0pt,
    partopsep=0pt
]
    \item \textbf{Logic Errors:} A logic error is a \emph{failure by action}. 
    This occurs when, given the current state, the agent reasons incorrectly and takes an action that results in failure. 
    For example, if in Frogger the agent can see that a car is about to move in front of it, but moves into that row anyway, causing a collision. 
    \item \textbf{Real-Time Errors:} A real-time error, by contrast, is a \emph{failure by inaction}. 
    This happens when the agent fails because the environment changed, and not because of an action taken by the agent.
    For example, if the agent is sitting in its lane while reasoning and a car hits the agent before it acts.
    Such an error is only possible in a real-time environment, where the environment updates without the agent acting.
\end{enumerate}

%
We use our criteria to analyse what our method and baselines learn during training, and how that translates to real-time deployment.
We compare to S1, since it is our best budget constrained baseline (beats Reactive in every setting) and to the Synchronous baseline, since AgileThinker is a deployment scaffold, not a trained method.

\begin{figure}[H]
\centering
\includegraphics[width=\columnwidth]{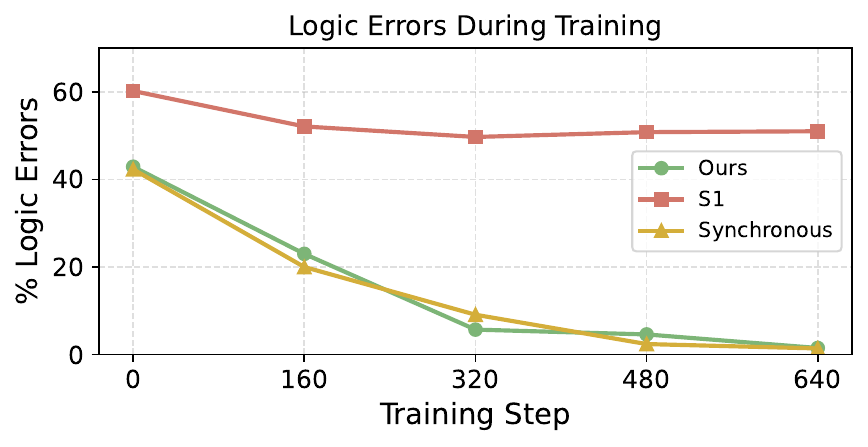}
\caption{Logic errors during training. Percentage of rollouts that end in a logic error over the course of training, for our method, S1, and the Synchronous baseline. Each checkpoint is evaluated on 256 held-out seeds with 8 rollouts each.}
\label{fig:logic-errors}
\end{figure}
\vspace{-3ex}

\textbf{Training analysis.} \Cref{fig:logic-errors} shows the percentage of trajectories that end in a logic error over training steps on in-distribution tasks in Frogger.
We see that S1 makes the most logic errors at the start of training, with 60\% of trajectories terminating in a logic error, compared to just 40\% for our method and Synchronous.
This is because S1 naively limits reasoning tokens per-action to $\Delta_{\mathcal{E}}$ to avoid real-time errors. 
By contrast, Synchronous has uncapped reasoning tokens, and at the start of training our method is effectively uncapped since the curriculum starts at very low time pressure.
S1 does learn to use its budget more efficiently, dropping its logic errors to 50\% by step 640, but both our method and Synchronous achieve near 0\% logic errors by the end of training.
In fact, our method matches the Synchronous error-rate throughout training, despite the effects of higher time pressure.
This suggests that our method learns to allocate reasoning in a non-naive way, such that it maintains the analytical power of a synchronous model even under high time pressure.

\begin{table}[H]
\centering
\footnotesize
\setlength{\tabcolsep}{4pt}
\begin{tabular*}{\columnwidth}{@{\extracolsep{\fill}} l l cccc @{}}
\toprule
& & \multicolumn{4}{c}{$\Delta_{\mathcal{E}}$} \\
\cmidrule(lr){3-6}
Method & Outcome & 400 & 200 & 100 & 50 \\
\midrule
Synchronous          & Real-time error & 50.8 & 92.0 & 99.7 & 100.0 \\
                     & Logic error     &  0.0 &  0.0 &  0.0 &   0.0 \\
                     & Won             & 49.2 &  8.0 &  0.3 &   0.0 \\
\midrule
S1                   & Real-time error &  0.0 &  0.0 &  0.0 &   0.0 \\
                     & Logic error     &  2.8 & 72.2 & 72.7 &  75.8 \\
                     & Won             & 97.2 & 27.8 & 27.3 &  24.2 \\
\midrule
Ours                 & Real-time error &  3.4 &  5.3 & 11.6 &  19.0 \\
                     & Logic error     &  2.0 &  2.9 &  0.1 &  11.7 \\
                     & Won             & 94.6 & 91.8 & 88.3 &  69.3 \\
\bottomrule
\end{tabular*}
\caption{Failure analysis. Percentage of evaluation episodes that end in a real-time error, end in a logic error, or are won, under each level of time pressure; smaller $\Delta_{\mathcal{E}}$ means more pressure. Computed over 256 held-out seeds, 8 rollouts each.}
\label{tab:failure}
\end{table}

\vspace{-5ex}

\textbf{Deployment analysis.} \Cref{tab:failure} breaks down final checkpoint performance by logic errors, real-time errors, and success on out-of-distribution tasks over a range of time pressures in Frogger.
As we saw previously, the Synchronous baseline learns to minimize logic errors. Correspondingly, it makes no logic errors in our evaluation.
Instead, the error rate is dominated by real-time errors, which occur in 100\% of trajectories under high ($\Delta_{\mathcal{E}}=50$) time pressure.
Budgeted methods have the opposite problem. By construction, S1 does not experience real-time errors since it must act every $\Delta_{\mathcal{E}}$ tokens.
Instead, as time pressure increases, S1 makes more logic errors as it is unable to reason accurately within its constrained budget.
Our method is the only one that learns to reduce both logic and real-time errors during training. 
Both error types increase with time pressure, but our method balances them, winning 45\% more often at the highest time pressure than the next-best baseline, S1.


\section{Related Work}
\label{sec:related}

\textbf{Real-time LLM agents.}
Previous work has sought to adopt LLM agents in real-time environments using scaffolded approaches.
These methods are inspired by dual-process theory, which posits that human cognition is the product of two distinct modes of processing: a fast, automatic mode (System 1) and a slow, deliberate mode (System 2).
In these scaffolds, System 1 is typically a small language model~\citep{liu2024llmpoweredhierarchicallanguageagent} or a vision transformer~\citep{cui2025openhelixshortsurveyempirical}, while System 2 is a tool-augmented LLM~\citep{christakopoulou2024agentsthinkingfastslow}.
However, these methods rely on a pre-defined ratio between slow and fast systems to control the amount of deliberate reasoning per-action.
By contrast, we are the first to employ reasoning training to natively instill this capability into our agent.
This helps to explain the difference in performance between our method and AgileThinker in \cref{exp:main}.

\textbf{Controlling reasoning length.}
LCPO~\citep{aggarwal2025l1controllinglongreasoning} and BudgetThinker~\citep{wen2025budgetthinkerempoweringbudgetawarellm} train models to obey explicit reasoning-length constraints, while S1~\citep{muennighoff2025s1simpletesttimescaling} controls its budget at inference-time.
These approaches make the time and cost of inference controllable. 
However, the budget must be specified a priori which does not transfer well to real-time environments, where the optimal amount of reasoning varies depending on the state.
Methods such as TALE, Thinkless, and AdapThink~\citep{han2024tale,fang2025thinkless,xu-etal-2026-adapthink} allocate compute based on properties such as problem difficulty, allowing harder problems to receive more reasoning than easier ones.
Our method is similar in that it learns to allocate reasoning according to problem difficulty, but unlike these methods it also accounts for time pressure, where extra reasoning may result in a worse answer.

\textbf{Learning transferable reasoning from games.}
Prior work shows that reinforcement learning in games can produce reasoning that transfers to held-out environments and benchmarks~\citep{cheng2024spag,spiral,xie2025vigal}.
These results demonstrate that game-based training can teach general reasoning strategies rather than only game-specific behavior.
We observe similar behavior in \cref{exp:main}, where RRT methods transfer well to out-of-distribution tasks. However, previous approaches show this in synchronous environments, while ours are real-time.

\section{Limitations and Conclusion}

We introduced real-time reasoning training (RRT), which teaches LLM agents when to keep reasoning and when to act, without a scaffold or a predefined reasoning budget.
In real-time games, our method stays robust as time pressure increases where baselines degrade, and it is the only method that learns to reduce both logic and real-time errors.
However, our curriculum requires control over the environment's update interval, which is unavailable in the physical world.
We also measure time in tokens, not wall-clock time, and evaluate a single 4B-parameter model in 2D games with discrete actions, so generalization beyond this setting remains unclear.
As our techniques are not game-specific, we hope RRT inspires time-aware agents in other domains.

\bibliography{main}
\bibliographystyle{icml2027}

\newpage
\appendix
\onecolumn
\section{Environments, Tasks and Rewards}
\label{apx:evaluation-tasks}
\Cref{tab:tasks} describes each task and its reward condition for both environments.

\begin{table}[h]
\centering
\footnotesize
\renewcommand{\arraystretch}{1.15}
\begin{tabular}{@{} l l >{\raggedright\arraybackslash}p{0.42\textwidth} >{\raggedright\arraybackslash}p{0.26\textwidth} @{}}
\toprule
Environment & Task type & Description & Reward condition \\
\midrule
Frogger & In-distribution & Cross a 10$\times$6 board from the bottom row to the top row while cars slide sideways along their rows and wrap around. & 1.0 for reaching the top row; 0.0 if hit by a car. \\
        & In-distribution & Same crossing, but three numbered checkpoint tiles on car rows must be visited in ascending order before reaching the top row. Stepping on a checkpoint out of order ends the game, as does reaching the top row. & Fraction of the four objectives completed (three checkpoints plus the top row). The top row only counts once all three checkpoints are visited. \\
        & In-distribution & The basic crossing rotated a quarter turn on a 6$\times$10 board: lanes are columns with cars sliding up or down, and the frog crosses from the rightmost column to the leftmost. & 1.0 for reaching the goal column; 0.0 if hit by a car. \\
        & Out-of-distribution & Round trip: reach the top row, then return to the exact starting cell on the bottom row while traffic keeps moving. Touching the top row does not end the game. & 1.0 for the full round trip; 0.5 for reaching the top but dying on the return; 0.0 for dying on the way up. \\
\midrule
Snake   & In-distribution & Snake on a 10$\times$8 board with 5 static boulders. Eat 5 food items, each respawning after it is eaten; hitting a wall, the snake's own body or a boulder ends the game. & 1.0 for eating all 5; otherwise food eaten / 5. \\
        & In-distribution & Same board with the roles swapped: eat 5 boulders and avoid 5 static apples. & 1.0 for eating all 5 boulders; otherwise boulders eaten / 5. \\
        & In-distribution & Five lettered objects (A--E) are placed at the start; a random 1--4 of them are deadly and the rest are collectible, as labelled in the state. Collect all the collectibles in any order; touching a deadly one ends the game. & Fraction of collectible objects eaten (the denominator varies per episode); 1.0 for collecting them all. \\
        & Out-of-distribution & Five numbered food items are placed at the start and must be eaten in ascending order (1 to 5). Only the lowest remaining number is safe; touching any higher number ends the game. & 1.0 for eating all 5 in order; otherwise items eaten / 5. \\
\bottomrule
\end{tabular}
\caption{Environments, tasks, and rewards. Each environment has three in-distribution tasks used for training and one out-of-distribution task held out for evaluation.}
\label{tab:tasks}
\end{table}


\end{document}